\documentclass[twocolumn]{article}

\newcommand{\pd}[2]{\frac{\partial #1}{\partial #2}}

\newcommand{\mb}{\mathbf}
\newcommand{\mc}{\mathcal}
\newcommand{\lp}{\left(}
\newcommand{\rp}{\right)}

\newcommand{\gij}[2]{T_{#1}\lp \mb x_{#2} | \mb x_{#1} \rp}
\newcommand{\gijr}[2]{\widetilde{T}_{#2}\lp \mb x_{#2} | \mb x_{#1} \rp}
\newcommand{\ygij}[2]{T_{\cup #1}\lp \mb y_{#2} | \mb y_{#1} \rp}
\newcommand{\ygijr}[2]{{\widetilde{T}}_{\cup #2}\lp \mb y_{#2} | \mb y_{#1} \rp}
\newcommand{\pit}[2]{\pi_{#1}\lp \mb x_{#2} \rp}
\newcommand{\Ept}[2]{E_{\pi_{#1}}\lp x_{#2} \rp}

\usepackage{times}
\usepackage{amsmath,amssymb}

\usepackage{color}

\usepackage{graphicx}
\usepackage{subfigure}

\usepackage{natbib}

\usepackage{algorithm}
\usepackage{algorithmic}

\usepackage{hyperref}

\begin{document}
\title{Hamiltonian Annealed Importance Sampling for partition function estimation}
\author{
Jascha Sohl-Dickstein and Benjamin J. Culpepper \\
Redwood Center for Theoretical Neuroscience\\
University of California, Berkeley\\
Berkeley, CA
\vspace{0.2in} \\
\vspace{-0.1in}
Redwood Technical Report
}

\maketitle

\begin{abstract}
We introduce an extension to annealed importance sampling that uses Hamiltonian dynamics to rapidly estimate normalization constants. We demonstrate this method by computing log likelihoods in directed and undirected probabilistic image models. We compare the performance of linear generative models with both Gaussian and Laplace priors, product of experts models with Laplace and Student's t experts, the mc-RBM, and a bilinear generative model. We provide code to compare additional models.
%
%
\end{abstract}

\section{Introduction}

We would like to use probabilistic models to assign probabilities to data. Unfortunately, this innocuous statement belies an important, difficult problem: many interesting distributions used widely across sciences cannot be analytically normalized.
Historically, the training of probabilistic models has been motivated in terms of maximizing the log probability of the data under the model or minimizing the KL divergence between the data and the model.  
However, for most models it is impossible to directly compute the log likelihood, due to the intractability of the  normalization constant, or partition function.
For this reason, performance is typically measured using a variety of diagnostic heuristics, not directly indicative of log likelihood.  For example, image models are often compared in terms of their synthesis, denoising, inpainting, and classification performance.
This inability to directly measure the log likelihood has made it difficult to consistently evaluate and compare models.

Recently, a growing number of researchers have given their attention to measures of likelihood in image models.
\cite{Salakhutdinov:2008p12541} use annealed importance sampling, and \cite{Murray:2009p12430} use a hybrid of annealed importance sampling and a Chib-style estimator to estimate the log likelihood of a variety of MNIST digits and natural image patches modeled using restricted Boltzmann machines and deep belief networks.
\cite{Bethge:2006p12046} measures the reduction in multi-information, or statistical redundancy, as images undergo various complete linear transformations.  \cite{Chandler:2007p12190} and \cite{Stephens:2008p12575} produce estimates of the entropy inherent in natural scenes, but do not address model evaluation. \cite{Karklin:2007p12497} uses kernel density estimates -- essentially, vector quantization -- to compare different image models, though that technique suffers from severe scaling problems except in specific contexts.
\cite{Zoran:2009p12559} compare the true log likelihoods of a number of image models, but restricts their analysis to the rare cases where the partition function can be solved analytically.

In this work, we merge two existing ideas -- annealed importance sampling and Hamiltonian dynamics -- into a single algorithm. To review, Annealed Importance Sampling (AIS) \cite{Neal:AIS} is a sequential Monte Carlo method \cite{Moral:2006p12473} which allows the partition function of a non-analytically-normalizable distribution to be estimated in an unbiased fashion. This is accomplished by starting at a distribution with a known normalization, and gradually transforming it into the distribution of interest through a chain of Markov transitions. Its practicality depends heavily on the chosen Markov transitions. Hamiltonian Monte Carlo (HMC) \cite{Neal:HMC} is a family of techniques for fast sampling in continuous state spaces, which work by extending the state space to include auxiliary momentum variables, and then simulating Hamiltonian dynamics from physics in order to traverse long iso-probability trajectories which rapidly explore the state space.

The key insight that makes our algorithm more efficient than previous methods is our adaptation of AIS to work with Hamiltonian dynamics. As in HMC, we extend the state space to include auxiliary momentum variables; however, we do this in such a way that the momenta change consistently through the intermediate AIS distributions, rather than resetting them at the beginning of each Markov transition.
To make the practical applications of this work clear, we use our method, Hamiltonian Annealed Importance Sampling (HAIS), to measure the log likelihood of holdout data under a variety of directed (generative) and undirected (analysis/feed-forward) probabilistic models of natural image patches.

The source code to reproduce our experiments is available.

\section{Estimating Log Likelihood}

\subsection{Importance Sampling}

Importance sampling \cite{Kahn:1953p12988} allows an unbiased estimate $\hat Z_p$ of the partition function (or normalization constant) $Z_p$ of a non-analytically-normalizable target distribution $p\lp \mb x \rp$ over $\mb x \in \mathbb R^M$,
\begin{align}
p\lp \mb x \rp &= \frac
{e^{-E_p\lp \mb x \rp}}
{Z_p}   \label{eq qx} \\
Z_p & = \int d\mb x\  
{e^{-E_p\lp \mb x \rp}}
,
\end{align}
to be calculated.  This is accomplished by averaging over samples $\mc S_q$ from a proposal distribution $q\lp \mb x \rp$,
\begin{align}
q\lp \mb x \rp &= \frac
{e^{-E_q\lp \mb x \rp}}
{Z_q} \\
Z_p & = \int d\mb x\  q\lp \mb x \rp
\frac
{e^{-E_p\lp \mb x \rp}}
{q\lp \mb x \rp}
                        \label{eq IS q ratio}
\\
\hat Z_p & = \frac{1}{\left| \mc S_q \right|}\sum_{x \in \mc S_q}
\frac
{e^{-E_p\lp \mb x \rp}}
{q\lp \mb x \rp}
                        \label{eq IS}
,
\end{align}
where $\left| \mc S_q \right|$ is the number of samples.  $q\lp \mb x \rp$ is chosen to be easy both to sample from and to evaluate exactly, and must have support everywhere that $p\lp \mb x \rp$ does.
Unfortunately, unless $q\lp \mb x \rp$ has significant mass everywhere $p\lp \mb x \rp$ does, it takes an impractically large number of samples from $q\lp \mb x \rp$ for $\hat Z_p$ to accurately approximate $Z_p$\footnote{
The expected variance of the estimate $\hat Z_p$ is given by an $\alpha$-divergence between $p\lp \mb x \rp$ and $q\lp \mb x \rp$, times a constant and plus an offset - see \cite{Minka:2005p12627}.
}.

\subsection{Annealed Importance Sampling}

Annealed importance sampling \cite{Neal:AIS} extends the state space $\mb x$ to a series of vectors, $\mb X = \left\{ \mb x_{1}, \mb x_{2} \ldots \mb x_{N} \right\}$, $\mb x_{n} \in \mathbb R^M$.  It then transforms the proposal distribution $q\lp \mb x \rp$ to a forward chain $Q\lp \mb X \rp$ over $\mb X$, by setting $q\lp \mb x \rp$ as the distribution over $\mb x_1$ and then multiplying by a series of Markov transition distributions,
\begin{align}
Q\lp \mb X \rp & = q\lp \mb x_{1} \rp \prod_{n=1}^{N-1} \gij{n}{n+1} 
,
\end{align}
where $\gij{n}{n+1}$ represents a \emph{forward} transition distribution from $\mb x_{n}$ to $\mb x_{n+1}$.
The target distribution $p\lp \mb x \rp$ is similarly transformed to become a reverse chain $P\lp \mb X \rp$, starting at $\mb x_{N}$, over $\mb X$,
\begin{align}
P\lp \mb X \rp & = \frac
    {e^{-E_p\lp \mb x_{N} \rp}}
    {Z_p}
    \prod_{n=1}^{N-1} \gijr{n+1}{n}
,
\end{align}
where $\gijr{n+1}{n}$ is a \emph{reverse} transition distribution from $\mb x_{n+1}$ to $\mb x_{n}$.  The transition distributions are, by definition, normalized (eg, $\int d\mb x\ _{n+1} \gij{n}{n+1} = 1$).

In a similar fashion to Equations \ref{eq IS q ratio} and \ref{eq IS}, samples $\mc S_Q$ from the forward proposal chain $Q\lp \mb X \rp$ can be used to estimate the partition function $Z_p$ (note that all integrals but the first in Equation \ref{eq Z_p int1} go to 1),
\begin{align}
Z_p & =     \int d\mb x_{N}\  {e^{-E_p\lp \mb x_{N} \rp}}
    \int d\mb x_{N-1}\  \gijr{N}{N-1}
    \nonumber \\ & \qquad \cdots
    \int d\mb x_{1}\  \gijr{2}{1} \label{eq Z_p int1} \\
& =     \int d\mb X\  Q\lp \mb X \rp \frac
        {e^{-E_p\lp \mb x_{N} \rp}}
        {Q\lp \mb X \rp}
    \gijr{N}{N-1}
    \nonumber \\ & \qquad
      \cdots \
    \gijr{2}{1}        \\
\hat Z_p & = \frac{1}{\left| \mc S_Q \right|}\sum_{X \in \mc S_Q}
    \frac
        {e^{-E_p\lp \mb x_{N} \rp}}
        {q\lp \mb x_{1} \rp}
    \frac
        {\gijr{2}{1}}
        {\gij{1}{2}}
    \nonumber \\ & \qquad
      \cdots
    \frac
        {\gijr{N}{N-1}}
        {\gij{N-1}{N}}
                        \label{eq AIS general}
.
\end{align}

In order to further define the transition distributions, Neal introduces intermediate distributions $\pi_{n}\lp \mb x \rp$ between $q\lp \mb x \rp$ and $p\lp \mb x \rp$,
\begin{align}
\pi_{n}\lp \mb x \rp & = \frac{e^{-E_{\pi_{n}}\lp x \rp
}}{Z_{\pi_{n}}} \\
E_{\pi_{n}}\lp \mb x \rp & = \lp1-\beta_{n}\rp E_q\lp \mb x \rp + \beta_{n} E_p\lp \mb x \rp
,
\end{align}
where the mixing fraction $\beta_{n} = \frac{n}{N}$ for all results reported here.
$\gij{n}{n+1}$ is then chosen to be any Markov chain transition for $\pi_{n}\lp \mb x \rp$, meaning that it leaves $\pi_{n}\lp \mb x \rp$ invariant
\begin{align}
T_{n} \circ \pi_{n} & = \pi_{n}
.
\end{align}
The reverse direction transition distribution $\gijr{n+1}{n}$ is set to the reversal of $\gij{n}{n+1}$,
\begin{align}
\gijr{n+1}{n} & = \gij{n}{n+1} \frac
{\pit{n}{n}}
{\pit{n}{n+1}}
.
\end{align}
Equation \ref{eq AIS general} thus reduces to
\begin{align}
\hat Z_p & = \frac{1}{\left| \mc S_Q \right|}\sum_{X \in \mc S_Q}
    \frac
        {e^{-E_p\lp \mb x_{N} \rp}}
        {q\lp \mb x_{1} \rp}
    \frac
        {\pit{1}{1}}
        {\pit{1}{2}}
    \nonumber \\ & \qquad
      \cdots
    \frac
        {\pit{N-1}{N-1}}
        {\pit{N-1}{N}} \\
& =     \frac{1}{\left| \mc S_Q \right|}\sum_{X \in \mc S_Q}
    \frac
        {e^{-E_p\lp \mb x_{N} \rp}}
        {q\lp \mb x_{1} \rp}
    \frac{e^{
        -\Ept{1}{1}
    }}
    {e^{
        - \Ept{1}{2}
    }}
    \nonumber \\ & \qquad
    \cdots
    \frac{e^{
        -\Ept{N-1}{N-1}
    }}
    {e^{
        - \Ept{N-1}{N}
    }}
                        \label{eq AIS}
.
\end{align}

If the number of intermediate distributions $N$ is large, and the transition distributions $\gij{n}{n+1}$ and $\gijr{n+1}{n}$ mix effectively, then the distributions over intermediate states $\mb x_{n}$ will be nearly identical to $\pit{n}{n}$ in both the forward and backward chains.  $P\lp \mb X \rp$ and $Q\lp \mb X \rp$ will then be extremely similar to one another, and the variance in the estimate $\hat Z_p$ will be extremely low\footnote{
There is a direct mapping between annealed importance sampling and the Jarzynski equality in non-equilibrium thermodynamics - see \cite{Jarzynski:1997p12846}.  It follows from this mapping, and the reversibility of quasistatic processes, that the variance in $\hat Z_p$ can be made to go to 0 if the transition from $q\lp \mb x_{1}\rp$ to $p\lp\mb x_{N}\rp$ is sufficiently gradual.
}.  If the transitions $\gij{n}{n+1}$ do a poor job mixing, then the marginal distributions over $\mb x_{n}$ under $P\lp \mb X \rp$ and $Q\lp \mb X \rp$ will look different from $\pit{n}{n}$.  The estimate $\hat Z_p$ will still be unbiased, but with a potentially larger variance.  Thus, to make AIS practical, it is important to choose Markov transitions $\gij{n}{n+1}$ for the intermediate distributions $\pi_{n}\lp \mb x \rp$ that mix quickly.

\subsection{Hamiltonian Annealed Importance Sampling}

Hamiltonian Monte Carlo \cite{Neal:HMC} uses an analogy to the physical dynamics of particles moving with momentum under the influence of an energy function to propose Markov chain transitions which rapidly explore the state space.  It does this by expanding the state space to include auxiliary momentum variables, and then simulating Hamiltonian dynamics to move long distances along iso-probability contours in the expanded state space. A similar technique is powerful in the context of annealed importance sampling. Additionally, by retaining the momenta variables across the intermediate distributions, significant momentum can build up as the proposal distribution is transformed into the target. This provides a mixing benefit that is unique to our formulation.

The state space $\mb X$ is first extended to $\mb Y = \left\{\mb y_{1}, \mb y_{2} \ldots \mb y_{N} \right\}$, $\mb y_{n} = \left\{ \mb x_{n}, \mb v_{n} \right\}$, where $\mb v_{n} \in \mathbb R^M$ consists of a momentum associated with each position $\mb x_{n}$.  The momenta associated with both the proposal and target distributions is taken to be unit norm isotropic gaussian.
The proposal and target distributions $q\lp \mb x \rp$ and $p\lp \mb x \rp$ are extended to corresponding distributions $q_\cup\lp \mb y \rp$ and $p_\cup\lp \mb y \rp$ over position and momentum $\mb y = \left\{ \mb x, \mb v \right\}$,
\begin{align}
p_\cup\lp \mb y \rp &= p\lp \mb x \rp \ \Phi\lp \mb v \rp = \frac
{e^{-E_{p_\cup}\lp \mb y \rp}}
{Z_{p_\cup}}
\\
q_\cup\lp \mb y \rp &= q\lp \mb x \rp \  \Phi\lp \mb v \rp = \frac
{e^{-E_{ q_\cup}\lp \mb y \rp}}
{Z_{q_\cup}}
\\
\displaybreak[0]
\Phi\lp \mb v \rp & = \frac
{e^{
    -\frac{1}{2} \mb v^T \mb v
}}
{\lp2\pi\rp^{\frac{M}{2}}}  
\\
\displaybreak[0]
E_{p_\cup}\lp \mb y \rp &=
E_p\lp \mb x \rp + \frac{1}{2} \mb v^T \mb v
\\
E_{q_\cup}\lp \mb y \rp &=
E_q\lp \mb x \rp + \frac{1}{2} \mb v^T \mb v
.
\end{align}
The remaining distributions are extended to cover both position and momentum in a nearly identical fashion: the forward and reverse chains $Q\lp \mb X \rp \rightarrow  Q_\cup\lp \mb Y \rp$, $P\lp \mb X \rp\rightarrow  P_\cup\lp \mb Y \rp$, the intermediate distributions and energy functions $\pi_{n}\lp \mb x \rp \rightarrow  \pi_{\cup \ n}\lp \mb y \rp$,  $E_{\pi_{n}}\lp \mb x \rp \rightarrow E_{\pi_{\cup \ n}}\lp \mb y \rp$,
\begin{align}
E_{\pi_{\cup \ n}}\lp \mb y \rp & = \lp1-\beta_{n}\rp E_{q_\cup}\lp \mb y \rp + \beta_{n} E_{p_\cup}\lp \mb y \rp
\\ &=
   \lp1-\beta_{n}\rp E_{q}\lp \mb x \rp + \beta_{n} E_{p}\lp \mb x \rp + \frac{1}{2} \mb v^T \mb v
,
\end{align}
and the forward and reverse Markov transition distributions $\gij{n}{n+1} \rightarrow \ygij{n}{n+1}$ and $\gijr{n+1}{n}\rightarrow \ygijr{n+1}{n}$.  Similarly, the samples $\mc S_{Q_\cup}$ now each have both position $\mb X$ and momentum $\mb V$, and are drawn from the forward chain described by $Q_\cup\lp \mb Y \rp$.

The annealed importance sampling estimate $\hat Z_p$ given in Equation \ref{eq AIS} remains {\em unchanged}, except for a replacement of $\mc S_{Q}$ with $\mc S_{Q_\cup}$ -- all the terms involving the momentum $\mb V$ conveniently cancel out, since the same momentum distribution $\Phi\lp \mb v \rp$ is used for the proposal $q_\cup\lp \mb y_{1} \rp$ and target $p_\cup\lp \mb y_{N} \rp$,
\begin{align}
\hat Z_p & =     \frac{1}{\left| \mc S_{Q_\cup} \right|}\sum_{Y \in \mc S_{Q_\cup}}
    \frac
       {e^{-E_p\lp \mb x_{N} \rp}    \Phi\lp \mb v_N \rp }
       {q\lp \mb x_{1} \rp                  \Phi\lp \mb v_1 \rp}
    \frac{e^{
       -\Ept{1}{1} + \frac{1}{2} \mb v_1^T \mb v_1
    }}
    {e^{
       - \Ept{1}{2} + \frac{1}{2} \mb v_2^T \mb v_2
    }}
    \nonumber \\ & \qquad
    \cdots
    \frac{e^{
       -\Ept{N-1}{N-1} + \frac{1}{2} \mb v_{N-1}^T \mb v_{N-1}
    }}
    {e^{
       - \Ept{N-1}{N} + \frac{1}{2} \mb v_N^T \mb v_N
    }}
\\ & =
\frac{1}{\left| \mc S_{Q_\cup} \right|}\sum_{Y \in \mc S_{Q_\cup}}
    \frac
       {e^{-E_p\lp \mb x_{N} \rp}}
       {q\lp \mb x_{1} \rp}
    \frac{e^{
       -\Ept{1}{1}
    }}
    {e^{
       - \Ept{1}{2}
    }}
    \nonumber \\ & \qquad\qquad
    \cdots
    \frac{e^{
       -\Ept{N-1}{N-1}
    }}
    {e^{
       - \Ept{N-1}{N}
    }}
                       \label{eq HAIS}
.
\end{align}
Thus, the momentum only matters when generating the samples $\mc S_{Q_\cup}$, by drawing from the initial proposal distribution $p_\cup\lp \mb y_{1} \rp$, and then applying the series of Markov transitions $\ygij{n}{n+1}$.

For the transition distributions, $\ygij{n}{n+1}$, we propose a new location by integrating Hamiltonian dynamics for a short time using a single leapfrog step, accept or reject the new location via Metropolis rules, and then partially corrupt the momentum.  That is, we generate a sample from $\ygij{n}{n+1}$ by following the procedure:
\begin{enumerate}
\item $\left\{ \mb x_{H}^0, \mb v_{H}^0 \right\} = \left\{ \mb x_{n}, \mb v_{n} \right\}$
\item leapfrog: \parbox[t]{1.6in}{$\mb x_{H}^\frac{1}{2} = \mb x_{H}^0 + \frac{\epsilon}{2} \mb v_{H}^0$ \vspace{3mm} \\
$\mb v_{H}^1 = \mb v_{H}^0 - \left. \epsilon \pd{E_{\pi_{n}}\lp \mb x \rp}{\mb x}\right|_{\mb x = \mb x_{H}^\frac{1}{2}}$ \\
$\mb x_{H}^1 = \mb x_{H}^\frac{1}{2} + \frac{\epsilon}{2} \mb v_{H}^1$} \vspace{2mm} \\
where the step size $\epsilon = 0.2$ for all experiments in this paper.
\item accept/reject: $\left\{ \mb x', \mb v' \right\} = \left\{ \mb x_{H}^1, -\mb v_{H}^1 \right\}$ with probability
$P_{accept} = \min\left[
    1,
    \frac{
     e^{
        -E_{\pi_{n}}\lp \mb x_{H}^1 \rp
        -\frac{1}{2}{\mb v_{H}^1}^T{\mb v_{H}^1}
    } }
    {
    e^{
        -E_{\pi_{n}}\lp \mb x_{H}^0 \rp
        -\frac{1}{2}{\mb v_{H}^0}^T{\mb v_{H}^0}
    } } \right]$, otherwise $\left\{ \mb x', \mb v' \right\} = \left\{ \mb x_{H}^0, \mb v_{H}^0 \right\}$
\item partial momentum refresh: $\tilde{\mb v}' = -\sqrt{1 - \gamma}\mb v' + \gamma \mb r$, where $r \sim \mc N\lp 0, \mb I \rp$, and $\gamma \in \lp 0, 1 \right]$ is chosen so as to randomize half the momentum power per unit simulation time \cite{anon_cvpr}.
\item $\mb y_{n+1} = \left\{ \mb x_{n+1}, \mb v_{n+1} \right\} = \left\{ \mb x', \tilde{\mb v}' \right\}$
\end{enumerate}
This combines the advantages of many intermediate distributions, which can lower the variance in the estimated $\hat Z_p$, with the improved mixing which occurs when momentum is maintained over many update steps.  For details on Hamiltonian Monte Carlo sampling techniques, and a discussion of why the specific steps above leave $\pi_{n}\lp \mb x \rp$ invariant, we recommend \cite{anon_cvpr,Neal:HMC}.

Some of the models discussed below have linear constraints on their state spaces.  These are dealt with by negating the momentum $\mb v$ and reflecting the position $\mb x$ across the constraint boundary every time a leapfrog halfstep violates the constraint.

\subsection{Log likelihood of analysis models}

Analysis models are defined for the purposes of this paper as those which have an easy to evaluate expression for $E_p\lp \mb x \rp$ when they are written in the form of Equation \ref{eq qx}.  The average log likelihood $\mc L$ of an analysis model $p\lp \mb x \rp$ over a set of testing data $\mc D$ is
\begin{align}
\mc L = \frac{1}{\left| \mc D \right|} \sum_{\mb x \in \mc D} \log p\lp \mb x \rp = -\frac{1}{\left| \mc D \right|} \sum_{\mb x \in \mc D} E_p\lp \mb x \rp - \log Z_p
\end{align}
where $\left| \mc D \right|$ is the number of samples in $\mc D$,
and the $Z_p$ in the second term can be directly estimated by Hamiltonian annealed importance sampling.

\subsection{Log likelihood of generative models}

Generative models are defined here to be those which have a joint distribution,
\begin{align}
p\lp \mb x, \mb a \rp
& =
p\lp \mb x | \mb a \rp p\lp \mb a \rp
=
\frac
{e^{-E_{x|a}\lp \mb x, \mb a \rp }}
{Z_{x|a}}
\frac
{e^{- E_{a}\lp \mb a \rp }}
{Z_a}
,
\end{align}
over visible variables $\mb x$ and auxiliary variables $\mb a \in \mathbb R^L$ which is easy to exactly evaluate and sample from, but for which the marginal distribution over the visible variables $p\lp \mb x \rp = \int d\mb a\ p\lp \mb x, \mb a \rp$ is intractable to compute.  The average log likelihood $\mc L$ of a model of this form over a testing set $\mc D$ is
\begin{align}
\mc L &= \frac{1}{\left| \mc D \right|} \sum_{\mb x \in \mc D} \log Z_{a|x} \\
Z_{a|x} &= \int d\mb a\ e^{    
-E_{x|a}\lp \mb x, \mb a \rp
-\log Z_{x|a}
- E_{a}\lp \mb a \rp
- \log Z_a
}
,
\end{align}
where each of the $Z_{a|x}$ can be estimated using HAIS.  Generative models take significantly longer to evaluate than analysis models, as a separate HAIS chain must be run for each test sample.

\section{Models} \label{sec models}

\begin{figure}[t]
\vskip 0.2in
\begin{center}
\includegraphics[width= 0.92\columnwidth]{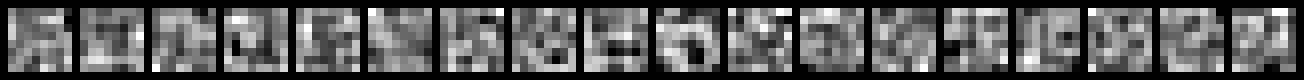}  (a) \\
\includegraphics[width= 0.92\columnwidth]{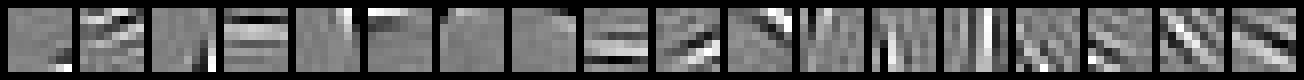}  (b) \\
\includegraphics[width= 0.92\columnwidth]{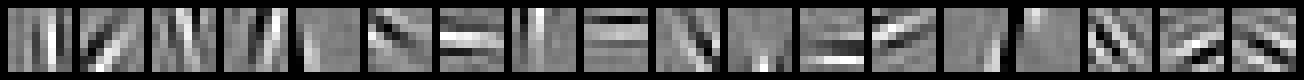}  (c) \\
\includegraphics[width= 0.92\columnwidth]{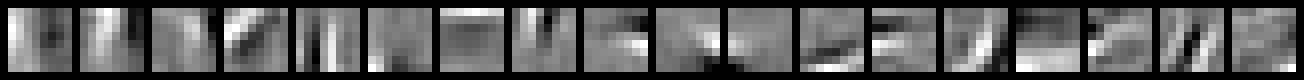} (d) \\
\includegraphics[width= 0.92\columnwidth]{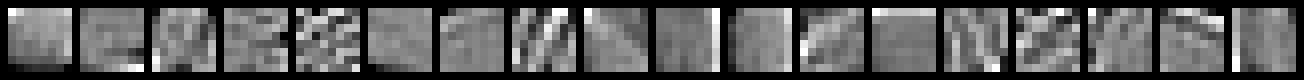} (e) \\
\includegraphics[width= 0.92\columnwidth]{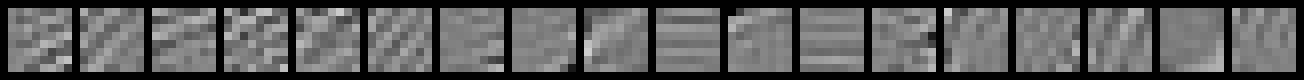} (f) \\
\includegraphics[width= 0.92\columnwidth]{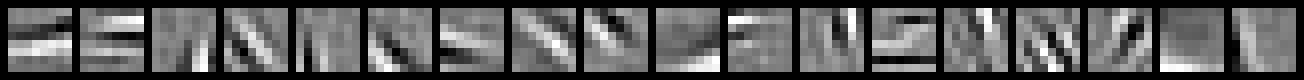} (g) \\
\includegraphics[width= 0.92\columnwidth]{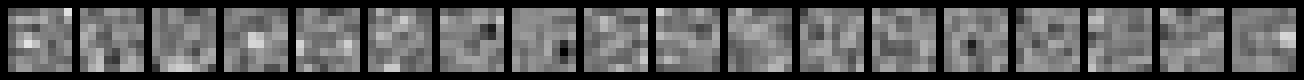} (h) \\
\includegraphics[width= 0.92\columnwidth]{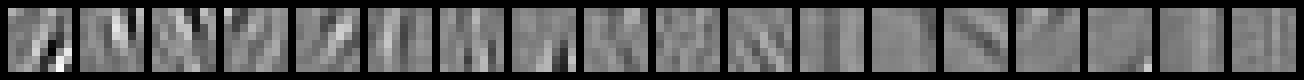} (i) \\
\caption{A subset of the basis functions and filters learned by each model. {\em (a)} Bases $\Phi$ for the linear generative model with Gaussian prior and {\em (b)} Laplace prior; {\em (c)} filters $\Phi$ for the product of experts model with Laplace experts, and {\em (d)} Student's t experts; {\em (e)} Bases $\Phi$ for the bilinear generative model and {\em (f)} the basis elements making up a single grouping from $\Psi$, ordered by and contrast modulated according to the strength of the corresponding $\Psi$ weight (decreasing from left to right); mcRBM {\em (g)} $C$ filters, {\em (h)} $W$ means, and {\em (i)} a single $P$ grouping, showing the pooled filters from $C$, ordered by and contrast modulated according to the strength of the corresponding $P$ weight (decreasing from left to right). }
\label{fig:bf}
\end{center}
\vskip -0.2in
\end{figure}

The probabilistic forms for all models whose log likelihood we evaluate are given below.  In all cases, $\mb x \in \mathbb R^M$ refers to the data vector.
\begin{enumerate}
\item linear generative:
\begin{align}
p\lp \mb x | \mb a \rp & = \frac{
    \exp\left[
        -\frac{1}{2\sigma_n^2} {\lp \mb x - \Phi \mb a \rp}^T {\lp \mb x - \Phi \mb a \rp}
    \right]
}{
    \lp 2 \pi \rp^{\frac{M}{2}} \sigma_n^M
}
\label{eq lin gen}
\end{align}
parameters: $\Phi \in \mathbb R^{M \times L}$\\
auxiliary variables: $\mb a \in \mathbb R^L$ \\
constant: $\sigma_n = 0.1$ \\
Linear generative models were tested with a two priors, as listed:
\begin{enumerate}
\item Gaussian prior: \label{sec lin gauss}
\begin{align}
p\lp \mb a \rp & = \frac{
    \exp\left[
        -\frac{1}{2} {\mb a}^T {\mb a}
    \right]
}
{
\lp 2 \pi \rp^{\frac{L}{2}}
}
\end{align}
\item Laplace prior \cite{Olshausen:1997p12767}:
\begin{align}
p\lp \mb a \rp & = \frac{
    \exp\left[
        -\left| \left| \mb a \right| \right|_1^1
    \right]
}
{
2
}
\end{align}
\end{enumerate}
\item bilinear generative \cite{anon_cvpr}: \label{mod bilin} The form is the same as for the linear generative model, but with the coefficients $\mb a$ decomposed into 2 multiplicative factors, 
\begin{align}
\mb a &= \lp\Theta \mb c\rp \odot \lp\Psi \mb d\rp
\\
p\lp \mb c \rp & = \frac{
    \exp\left[
        -\left| \left| \mb c \right| \right|_1^1
    \right]
}
{
2
} \\
p\lp \mb d \rp & =
    \exp\left[
        -\left| \left| \mb d \right| \right|_1^1
    \right]
,
\end{align}
where $\odot$ indicates element-wise multiplication. \\
parameters: $\Phi \in \mathbb R^{M \times L}$, $\Theta \in \mathbb R^{L \times K_c}$, $\Psi \in \mathbb R^{L \times K_d}$ \\
auxiliary variables: $\mb c \in \mathbb R^{K_c}$, $\mb d \in \mathbb R_+^{K_d}$
\item product of experts \cite{Hinton2002}: This is the analysis model analogue of the linear generative model,  \label{sec POE}
\begin{align}
p\lp \mb x \rp & = \frac{1}{Z_{POE}}
\prod_{l=1}^L \exp\lp -E_{POE}\lp \Phi_l \mb x; \lambda_l  \rp \rp
.
\end{align}
parameters: $\Phi \in \mathbb R^{L \times M}$, $\lambda \in \mathbb R_+^L$,\\
Product of experts models were tested with two experts, as listed:
\begin{enumerate}
\item Laplace expert:
\begin{align}
E_{POE}\lp u; \lambda_l  \rp = \lambda_l \left| u \right|
\end{align}
(changing $\lambda_l$ is equivalent to changing the length of the row $\Phi_l$, so it is fixed to $\lambda_l = 1$)
\item Student's t expert:
\begin{align}
E_{POE}\lp u; \lambda_l \rp = \lambda_l \log\lp 1 + u^2 \rp
\end{align}
\end{enumerate}
\item Mean and covariance restricted Boltzmann machine (mcRBM) \cite{Ranzato:2010p12217}:  This is an analysis model analogue of the bilinear generative model.  The exact marginal energy function $E_{mcR}$ is taken from the released code rather than the paper.
\begin{align}
p\lp \mb x \rp & = \frac{
 \exp\left[ - E_{mcR}\lp \mb x \rp \right]
 }{Z_{mcR}} \\
E_{mcR}\lp \mb x \rp & = -\sum_{k=1}^K  \log \left[ 1 + e^{
                         \frac{1}{2}
                             \sum_{l=1}^L P_{lk} \frac{
                                 \left( \mb C_l \mb x \right)^2
                             }{
                                 \left| \left| \mb x \right|\right|_2^2 + \frac{1}{2}
                             }
                             + b^c_k`
                         } \right]
 \nonumber \\ & \qquad
                      -\sum_{j=1}^J \log \left[ 1 + e^{
                         \mb W_j \mb x + b^m_j
                         } \right]
 \nonumber \\ & \qquad
                     +\frac{1}{2\sigma^2} \mb x^T \mb x
                     -\mb x^T \mb b^v
\end{align}
parameters: $P \in \mathbb R^{L \times K}$, $C \in \mathbb R^{L \times M}$, $W \in \mathbb R^{J \times M}$, $b^m \in \mathbb R^J$, $b^c \in \mathbb R^K$, $b^v \in \mathbb R^K$, $\sigma \in \mathbb R$
\end{enumerate}

\section{Training}

All models were trained on 10,000 $16x16$ pixel image patches taken at random from 4,112 linearized images of natural scenes from the van Hateren dataset \cite{Hateren:1998p12895}.  The extracted image patches were first logged, and then mean subtracted.  They were then projected onto the top $M$ PCA components, and whitened by rescaling each dimension to unit norm.

All generative models were trained using Expectation Maximization over the full training set, with a Hamiltonian Monte Carlo algorithm used during the expectation step to maintain samples from the posterior distribution.  See \cite{anon_cvpr} for details.  All analysis models were trained using LBFGS on the minimum probability flow learning objective function for the full training set, with a transition function $\Gamma$ based on Hamiltonian dynamics.  See \cite{MPF_ICML} for details.  
No regularization or decay terms were required on any of the model parameters. 

\section{Results}
\begin{figure}[t]
\vskip 0.2in
\begin{center}
\centerline{\includegraphics[width= 0.9\columnwidth]{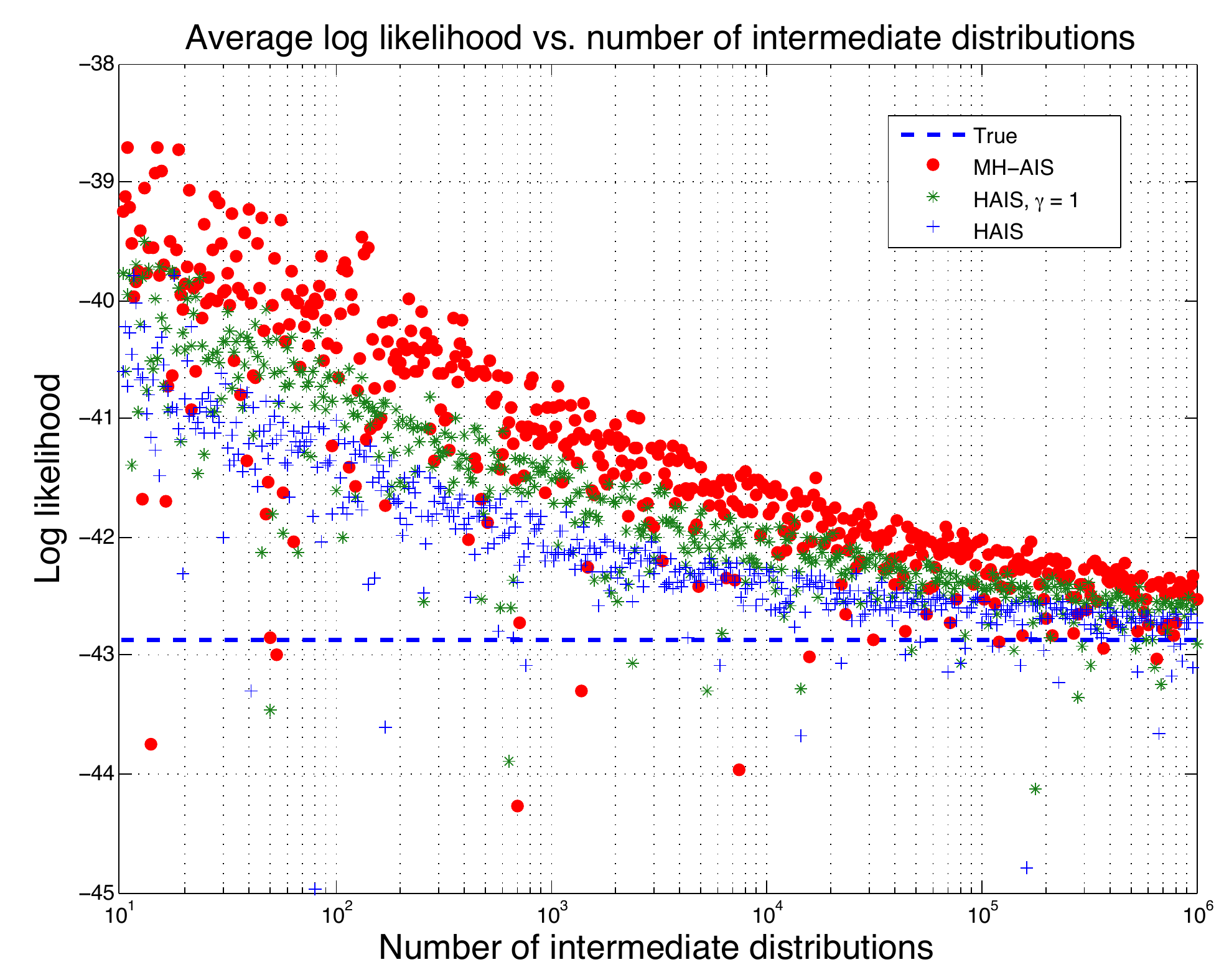}}
\caption{Comparison of HAIS with alternate AIS algorithms in a complete ($M=L=36$) POE Student's t model.  The scatter plot shows estimated log likelihoods under the test data for the POE model for different numbers of intermediate distributions $N$. The \textcolor{blue}{blue crosses} indicate HAIS.  The \textcolor{green}{green stars} indicate AIS with a single Hamiltonian dynamics leapfrog step per distribution, but no continuity of momentum. The \textcolor{red}{red dots} indicate AIS with a Gaussian proposal distribution.  The dashed \textcolor{blue}{blue line} indicates the true log likelihood of the minimum probability flow trained model.  
This product of Student's t distribution is extremely difficult to normalize numerically, as many of its moments are infinite.
}
\label{fig:converge-poe-studentt}
\end{center}
\vskip -0.2in
\end{figure}

100 images from the van Hateren dataset were chosen at random and reserved as a test set for evaluation of log likelihood.  The test data was preprocessed in an identical fashion to the training data.  Unless otherwise noted, log likelihood is estimated on the same set of 100 patches drawn from the test images, using Hamiltonian annealed importance sampling with $N=100,000$ intermediate distributions, and 200 particles.  This procedure takes about 170 seconds for the 36 PCA component analysis models tested below.  The generative models take approximately 4 hours, because models with unmarginalized auxiliary variables require one full HAIS run for each test datapoint.

\subsection{Validating Hamiltonian annealed importance sampling}

\begin{figure}[t]
\vskip 0.2in
\begin{center}
\centerline{\includegraphics[width= 0.9\columnwidth]{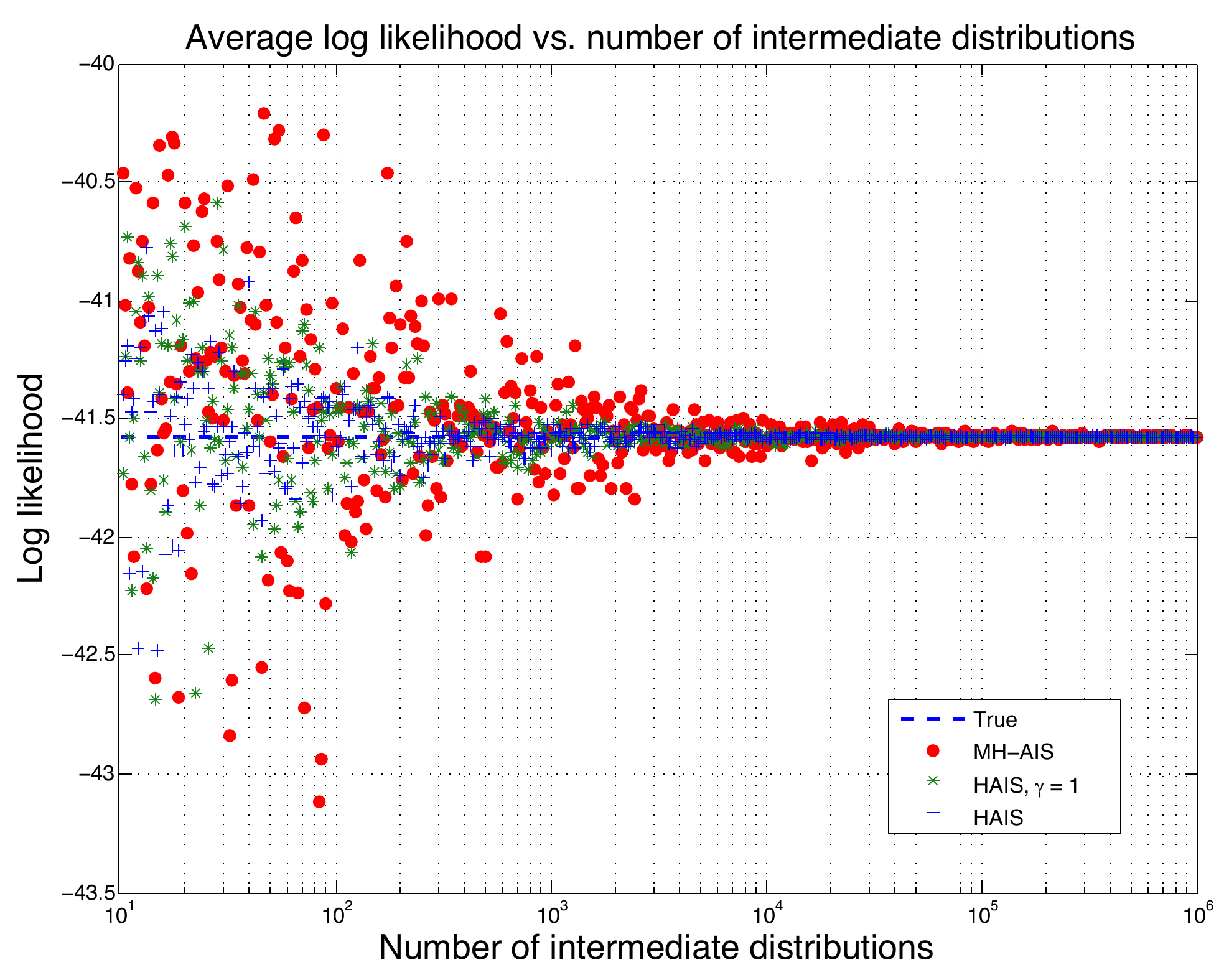}}
\caption{Comparison of HAIS with alternate AIS algorithms in a complete ($M=L=36$) POE Laplace model.  Format as in Figure \ref{fig:converge-poe-studentt}, but for a Laplace expert.
}
\label{fig:converge-poe-laplace}
\end{center}
\vskip -0.2in
\end{figure}

\begin{figure}[t]
\vskip 0.2in
\begin{center}
\centerline{\includegraphics[width= 0.9\columnwidth]{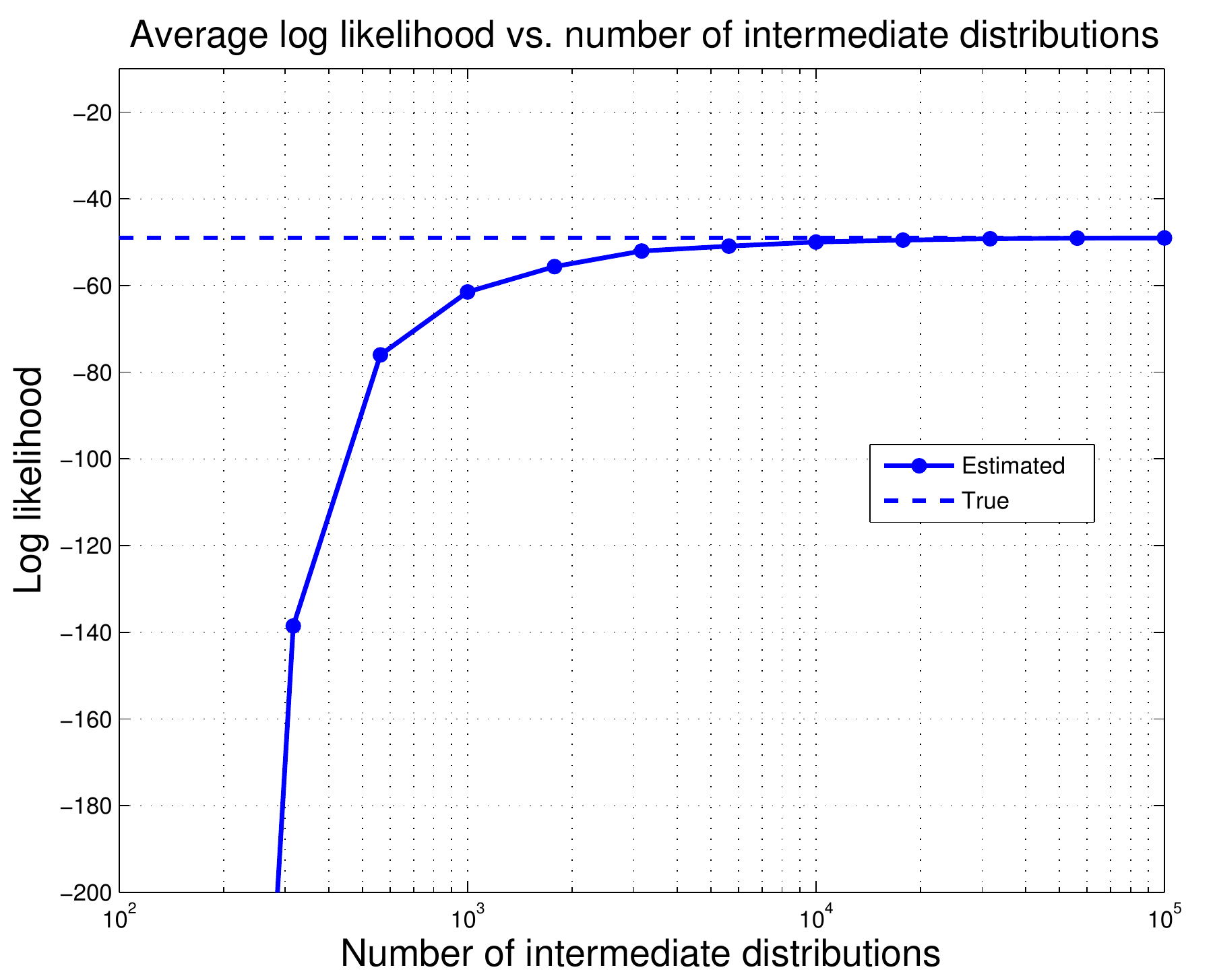}}
\caption{Convergence of HAIS for the linear generative model with a Gaussian prior. The dashed blue line indicates the true log likelihood of the test data under the model. The solid blue line indicates the HAIS estimated log likelihood of the test data for different numbers of intermediate distributions $N$.}
\label{fig:converge-linear-gauss}
\end{center}
\vskip -0.2in
\end{figure}

\begin{figure}[t]
\vskip 0.2in
\begin{center}
\centerline{\includegraphics[width= 0.9\columnwidth]{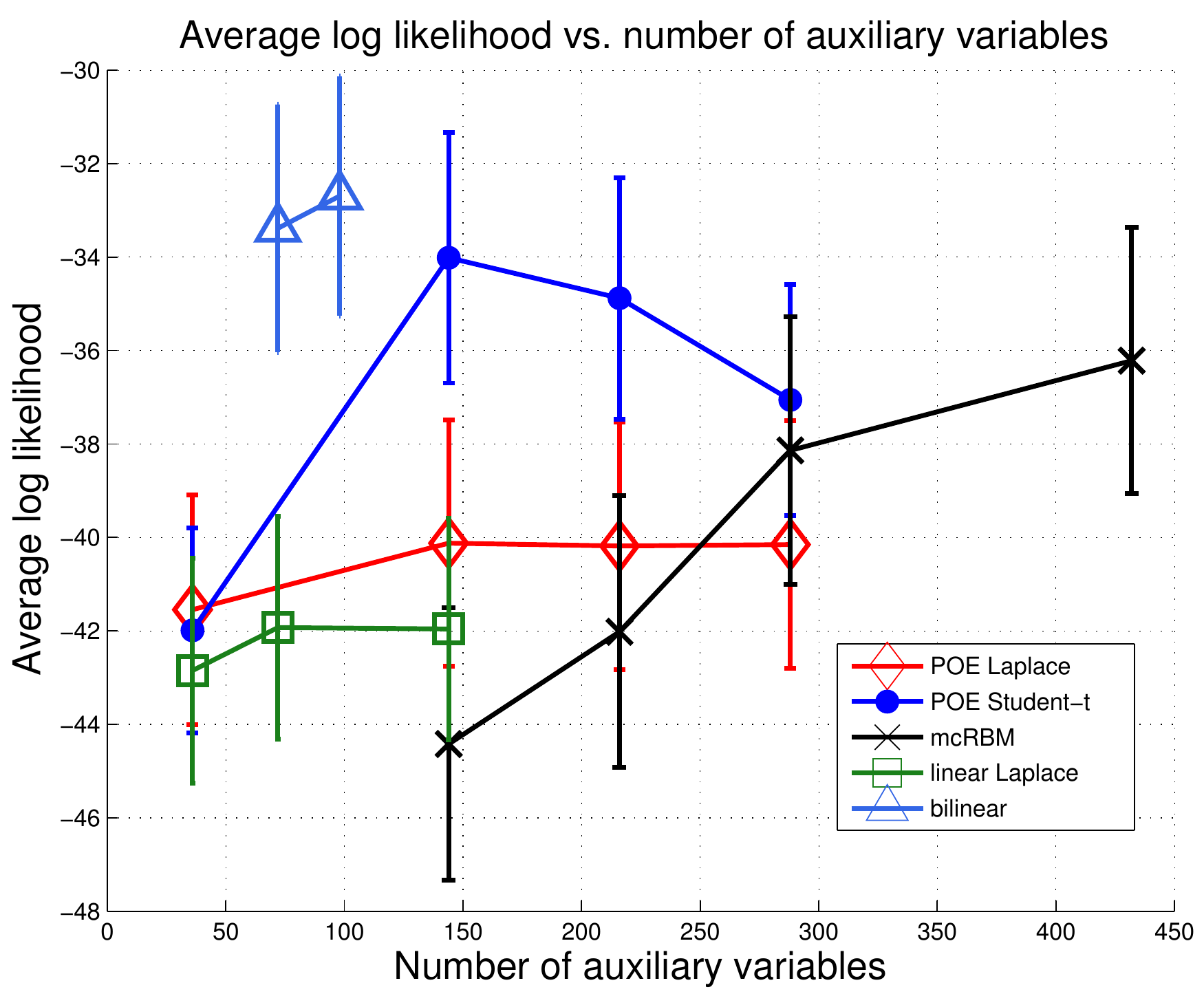}}
\caption{Increasing the number of auxiliary variables in a model increases the likelihood it assigns to the test data until it saturates, or overfits. }
\label{fig:fit}
\end{center}
\vskip -0.2in
\end{figure}

\begin{table}[t]
\caption{Average log likelihood for the test data under each of the models. The model `size' column denotes the number of experts in the POE models, the sum of the mean and covariance units for the mcRBM, and the total number of latent variables in the generative models.}
\label{table:model comparison}
\vskip 0.15in
\begin{center}
\begin{small}
\begin{sc}
\begin{tabular}{lrl}
\hline
Model & Size & Log Likelihood \\
\hline
Lin. generative, Gaussian & 36 & -49.15$\pm$ 2.31 \\
Lin. generative, Laplace & 36 & -42.85$\pm$ 2.41 \\
POE, Laplace experts & 144 & -41.54$\pm$ 2.46 \\
mcRBM & 432 & -36.01$\pm$ 2.57 \\
POE, Student's t experts & 144 &  -34.01$\pm$ 2.68 \\
Bilinear generative & 98 & -32.69$\pm$ 2.56 \\
\hline
\end{tabular}
\end{sc}
\end{small}
\end{center}
\vskip -0.1in
\end{table}

The log likelihood of the test data can be analytically computed for three of the models outlined above: linear generative with Gaussian prior (Section \ref{sec models}, model \ref{sec lin gauss}), and product of experts with a complete representation ($M = L$) for both Laplace and Student's t experts (Section \ref{sec models}, model \ref{sec POE}).  Figures \ref{fig:converge-poe-studentt}, \ref{fig:converge-poe-laplace} and \ref{fig:converge-linear-gauss} show the convergence of Hamiltonian annealed importance sampling, with 200 particles, for each of these three models as a function of the number $N$ of intermediate distributions.  Note that the Student's t expert is a pathological case for sampling based techniques, as for several of the learned $\lambda_l$ even the first moment of the Student's t-distribution was infinite.

Additionally, for all of the generative models, if $\mb \Phi = \mb 0$ then the statistical model reduces to,
\begin{align}
p\lp \mb x | \mb a \rp & = \frac{
    \exp\left[
        -\frac{1}{2\sigma_n^2} {\mb x}^T {\mb x}
    \right]
}{
    \lp 2 \pi \rp^{\frac{M}{2}} \sigma_n^M
} \, ,
\label{eq lin gen}
\end{align}
and the log likelihood $\mathcal L$ has a simple form that can be used to directly verify the estimate computed via HAIS.  We performed this sanity check on all generative models, and found the HAIS estimated log likelihood converged to the true log likelihood in all cases.

\subsection{Speed of convergence}

In order to demonstrate the improved performance of HAIS, we compare against two alternate AIS learning methods.  First, we compare to AIS with transition distributions $\gij{n}{n+1}$ consisting of a Gaussian ($\sigma_{diffusion} = 0.1$) proposal distribution and Metropolis-Hastings rejection rules.  Second, we compare to AIS with a single Hamiltonian leapfrog step per intermediate distribution $\pit{n}{n}$, and unit norm isotropic Gaussian momentum.  Unlike in HAIS however, in this case we randomize the momenta before each update step, rather than allowing them to remain consistent across intermediate transitions.  As can be seen in Figures \ref{fig:converge-poe-studentt} and \ref{fig:converge-poe-laplace}, HAIS requires fewer intermediate distributions by an order of magnitude or more.

\subsection{Model size}

By training models of different sizes and then using HAIS to compute their likelihood, we are able to explore how each model behaves in this regard, and find that three have somewhat different characteristics, shown in Figure~\ref{fig:fit}. The POE model with a Laplace expert has relatively poor performance and we have no evidence that it is able to overfit the training data; in fact, due to the relatively weak sparsity of the Laplace prior, we tend to think the only thing it can learn is oriented, band-pass functions that more finely tile the space of orientation and frequency. In contrast, the Student-t expert model rises quickly to a high level of performance, then overfits dramatically. Surprisingly, the mcRBM performs poorly with a number of auxiliary variables that is comparable to the best performing POE model. One explanation for this is that we are testing it in a regime where the major structures designed into the model are not of great benefit. That is, the mcRBM is primarily good at capturing long range image structures, which are not sufficiently present in our data because we use only 36 PCA components. Although for computational reasons we do not yet have evidence that the mcRBM can overfit our dataset, it likely does have that power. We expect that it will fare better against other models as we scale up to more sizeable images. Finally, we are excited by the superior performance of the bilinear generative model, which outperforms all other models with only a small number of auxiliary variables. We suspect this is mainly due to the high degree of flexibility of the sparse prior, whose parameters (through $\Theta$ and $\Psi$) are learned from the data. The fact that for a comparable number of ``hidden units'' it outperforms the mcRBM, which can be thought of as the bilinear generative model's `analysis counterpart', highlights the power of this model.

\subsection{Comparing model classes}

As illustrated in Table \ref{table:model comparison}, we used HAIS to compute the log likelihood of the test data under each of the image models in Section \ref{sec models}.  The model sizes are indicated in the table -- for both POE models and the mcRBM they were chosen from the best performing datapoints in Figure \ref{fig:fit}.  In linear models, the use of sparse priors or experts leads to a large ($> 6\ nat$) increase in the log likelihood over a Gaussian model.  The choice of sparse prior was similarly important, with the POE model with Student's t experts performing more than $7\ nats$ better than the POE or generative model with Laplace prior or expert.  Although previous work \cite{Ranzato:2010p12217,anon_cvpr} has suggested bilinear models outperform their linear counterparts, our experiments show the Student's t POE performing within the noise of the more complex models.  One explanation is the relatively small dimensionality (36 PCA components) of the data -- the advantage of bilinear models over linear is expected to increase with dimensionality. 
Another is that Student's t POE models are in fact better than previously believed. Further investigation is underway.  The surprising performance of the Student's t POE, however, highlights the power and usefulness of being able to directly compare the log likelihoods of probabilistic models.
%

\section{Conclusion}

By improving upon the available methods for partition function estimation, we have made it possible to directly compare large probabilistic models in terms of the likelihoods they assign to data. This is a fundamental measure of the quality of a model -- especially a model trained in terms of log likelihood -- and one which is frequently neglected due to practical and computational limitations.  It is our hope that the Hamiltonian annealed importance sampling technique presented here will lead to better and more relevant empirical comparisons between models.

\bibliography{report}
\bibliographystyle{icml2011}

\end{document}